\documentclass{jingdong}

\usepackage{lmodern}
\renewcommand{\sfdefault}{jdlangzhengti}
\usepackage{amsmath,amsfonts,bm}

\def\eqref#1{equation~\ref{#1}}

\def\1{\bm{1}}

\DeclareMathAlphabet{\mathsfit}{\encodingdefault}{\sfdefault}{m}{sl}
\SetMathAlphabet{\mathsfit}{bold}{\encodingdefault}{\sfdefault}{bx}{n}

\usepackage{amsmath}
\usepackage{amssymb}

\usepackage{booktabs}
\usepackage{multirow}
\usepackage{tabularx}
\usepackage{arydshln}
\usepackage{xcolor}
\usepackage{colortbl}

\usepackage{graphicx}
\usepackage{adjustbox}

\usepackage{algorithm}
\usepackage{algorithmic}

\usepackage{xspace}

\usepackage{enumitem}

\newenvironment{rqanswer}[1]{%
  \par\noindent\textbf{Answer to #1.}\ \ignorespaces
}{%
  \par
}

\hypersetup{hypertexnames=false}
\AtBeginEnvironment{quote}{\sloppy}

\title{Avoiding What Your Teacher Already Knows: Entity Familiarity for Deep-Search Distillation}
\author[]{Yifei~Liu}
\author[]{Zhixiang~Liang}
\author[]{Wentao~Tan}
\author[]{Yicheng~Wang}
\author[]{Jiaqi~Wang}
\author[]{Nan~Duan}
\author[\dagger]{Qiong~Cao}
\affiliation[]{JD.COM}
\contribution[\dagger]{~Corresponding author.}
\abstract{
Deep-search agents are commonly trained by distilling tool-use trajectories from stronger teacher models. However, structurally complex questions may remain relatively easy for teachers familiar with key entities in the underlying entity graph, resulting in lower search demand and weaker supervision. We refer to this teacher-specific variation in question difficulty as \textit{teacher-relative difficulty} and investigate entity familiarity as an early predictor. We first introduce Evidence-Verified Entity Familiarity (\textbf{EVF}), which estimates familiarity by verifying a teacher-generated factual brief against online evidence. Across multiple teachers, EVF is significantly correlated with teacher-relative difficulty with AUCs of 0.60--0.70. However, its high cost makes it impractical for constructing entity graphs at scale. To address this limitation, we propose Self-Reported Entity Familiarity (\textbf{SEF}), a lightweight gating signal that directly asks a teacher whether it is familiar with an entity before admitting that entity into the graph. Across multiple teachers, trajectories from SEF-gated synthesis exhibit greater observed search demand and yield higher downstream student accuracy than those from ungated synthesis. Because familiarity is teacher specific, we further study cross-teacher reuse. We screen graphs collected for other teachers using Kimi-K3's SEF judgments and use the retained graphs to produce 5,411 additional trajectories. Combined with 2,200 trajectories from newly constructed, SEF-gated graphs, these data improve Qwen3.8-27B's accuracy from 44.10\% to 66.41\% on a BrowseComp subset, and achieving state-of-the-art performance among models of comparable size on BrowseComp, SEAL-0, HLE-text, and xbench. Overall, SEF provides a low-cost way to synthesize more informative supervision for deep-search distillation.
}
\correspondence{\email{caoqiong1@jd.com}}
\checkdata[Code and Data]{https://feifeifeiliu.github.io/SEF/}

\begin{document}
\maketitle

\section{Introduction}
\label{sec:introduction}

Deep search requires a language model to interact with the open web, retrieving, reading, and integrating evidence before committing to an answer. Since its introduction in 2025, BrowseComp, a challenging benchmark for deep search, has rapidly become a standard evaluation for frontier models with web-search capabilities, with the state-of-the-art Kimi-K3 achieving 91.2\% accuracy \citep{browsecomp,tongyi_dr,deepseek_v32,kimik3}. Two common settings call for transferring this capability: distillation into a much smaller student, which reduces inference cost and enables local deployment \citep{ondevice_survey}, and the synthesis of training data for subsequent model generations \citep{phi4,llama3}. In both settings, a synthesis system constructs question--answer pairs, and the teacher solves the questions with search tools to produce trajectories as supervision \citep{websailor,opendatasynthesis}. For this supervision to be effective, the trajectories should exhibit substantive information acquisition and search reasoning.

A natural way to elicit stronger search behavior is to synthesize more difficult questions. Commonly used graph-based question--answer synthesis can increase the number of hops, constraints, and compositional steps, but these properties do not necessarily imply high search demand for a particular teacher. If the teacher already knows the relevant entities, relations, or likely answers, it may narrow the search space and recognize supporting evidence with little exploration. The resulting trajectories can therefore contain fewer meaningful search decisions and less information acquisition, providing weaker supervision for a student that may not share the teacher's prior knowledge. Indeed, recent work reports that models can answer some BrowseComp questions without tool access \citep{livebrowsecomp}. We use \emph{teacher-relative difficulty} to describe how much information a particular teacher must still obtain externally to solve a question. We use question-level closed-book correctness as a post-hoc measure of this quantity: if a teacher can answer a completed question without search tools, the question has low teacher-relative difficulty for that teacher. However, this signal is available only after the costly question synthesis, thus motivating a low-cost predictor that can be applied during question synthesis.

We hypothesize that \emph{entities less familiar to the teacher are more likely to yield questions with high teacher-relative difficulty}. We first test this hypothesis with \emph{Evidence-Verified Entity Familiarity} (EVF), which estimates familiarity by asking the teacher to produce a factual brief about an entity and verifying its claims against online evidence. Across three teacher-specific question pools, EVF measured on seed entities predicts the teacher's closed-book correctness with AUCs of 0.60--0.70; lower scores correspond to questions less likely to be answered correctly without tools. EVF, however, requires a factual brief, web search, and claim verification for each entity, making repeated screening during graph construction costly. We observe that within the low-EVF region, teachers often report being unfamiliar with the corresponding seed entity. This motivates examining whether a direct familiarity judgment can serve as a cheaper screening signal.

We therefore introduce \emph{Self-Reported Entity Familiarity} (SEF), which reduces familiarity estimation to a single tool-free binary query asking whether the teacher is familiar with an entity. Unlike EVF, SEF requires neither search nor evidence verification, making it inexpensive enough to apply during graph construction. We apply SEF at two entity-admission points of the graph-based synthesis pipeline: seed selection and next-hop expansion. \emph{SEF-gated synthesis} admits only entities that the teacher reports as unfamiliar, whereas \emph{ungated synthesis} omits this screening; all other stages are held fixed. Relative to ungated synthesis, trajectories produced by SEF-gated synthesis for GLM-5.1~\citep{glm51}, DeepSeek-V4-Flash~\citep{deepseek_v4}, and Kimi-K3 contain more search turns, reasoning tokens, and tool calls, together with lower tool-assisted solve accuracy, a pattern consistent with higher observed search demand. In the corresponding equal-sized distillation comparisons, students trained on SEF-gated trajectories show gains ranging from 6.15 to 8.46 percentage points over those trained on ungated trajectories.

Because SEF is teacher-specific, a natural question is whether data collected for one teacher can still benefit another. On a shared pool of 5,995 entities, pairwise SEF agreement across the three teachers is 84.4--89.9\%. Thus, different teachers share substantial overlap in entity familiarity, but their judgments are not identical. We therefore reuse the collected entity graphs and obtain new SEF judgments for the target teacher. We then resynthesize the question--answer pairs and rerun the target teacher to collect new search trajectories. With Kimi-K3 as the target, this process yields 5,411 trajectories from graphs collected for the other teachers. Combined with 2,200 trajectories generated through Kimi-K3 SEF-gated synthesis, the resulting 7,611 trajectories improve the accuracy of Qwen3.8-27B~\citep{qwen38_27b} from 44.10\% to 66.41\% on the evaluated BrowseComp subset.

Our contributions are threefold:
\begin{itemize}[leftmargin=*]
\item We formulate deep-search data difficulty in teacher-relative terms and show that entity familiarity is predictive of whether a teacher can answer a question without search. We study this relationship through both evidence-verified familiarity (EVF) and the much cheaper self-reported entity familiarity (SEF).

\item We introduce SEF-gated entity admission as an early intervention in graph-based question--answer synthesis. Across three teacher settings, trajectories from SEF-gated synthesis show higher observed search demand than those from ungated synthesis, and the corresponding students score higher by $+8.20$, $+6.15$, and $+8.46$ percentage points in the evaluated comparisons.

\item We study how teacher-specific familiarity affects data reuse across teachers. By reusing previously collected entity graphs to expand Kimi-K3-specific training data, we train a Qwen3.8-27B student that achieves 66.41\% accuracy on the BrowseComp subset.
\end{itemize}

\section{Problem Setup and Study Design}
\label{sec:study-setup}

We study teacher-specific entity familiarity in the setting of deep-search distillation. Our experiments build on an existing graph-based question--answer synthesis pipeline. This section introduces the distillation setting, describes the graph-based synthesis pipeline and our intervention points, and outlines the research questions investigated in the following sections.

\subsection{Deep-Search Distillation Setting}
\label{sec:deep-search-setting}

We consider distillation from a search-enabled teacher model $M_T$ to a student model $M_S$. Following ReAct-style search agents such as OpenSeeker~\citep{openseeker}, the teacher solves a question $q$ by producing a trajectory
\begin{equation}
    \tau_T(q)
    =
    \bigl[(r_1,a_1,o_1),\ldots,(r_m,a_m,o_m),\hat{y}\bigr],
    \label{eq:search-trajectory}
\end{equation}
where $r_t$, $a_t$, and $o_t$ denote reasoning, tool action, and observation at step $t$, respectively, and $\hat{y}$ is the final answer. A synthesis pipeline produces
\begin{equation}
    \mathcal{D}_T
    =
    \left\{(q_i,\tau_T(q_i))\right\}_{i=1}^{N},
    \label{eq:teacher-distillation-data}
\end{equation}
which is used to train the student with supervised fine-tuning on the teacher trajectories.

We distinguish structural complexity from teacher-relative difficulty. Structural complexity concerns properties such as the number of hops or constraints in a question, whereas teacher-relative difficulty concerns how much information a particular teacher must still acquire externally to solve it. A structurally complex question may therefore remain easy for a teacher already familiar with its key entities. In this work, teacher-relative difficulty is treated as a latent task property and studied through observable proxies such as closed-book answerability and search behavior. Note that we define closed-book correctness as whether the model answers a question correctly without tools. 

\subsection{Graph-Based Synthesis Pipeline}
\label{sec:synthesis-pipeline}

We use a graph-based synthesis pipeline adapted from OffSeeker~\citep{offseeker}. The synthesis process begins by retrieving webpages from diverse topics and extracting seed entities. Let $e$ denote an entity encountered during synthesis. Starting from a seed entity $e_0$, a tool-augmented agent searches the web, opens relevant pages, and collects factual properties of the current entity together with evidence-supported relations to other entities. The agent proposes related entities as candidates for the next hop, selects one, and repeats the exploration for multiple hops. This process constructs an evidence graph and a corresponding transition path,
\begin{equation}
    G=(V,R), \qquad
    P=(e_0,\ldots,e_h), \quad
    e_j\in V,\quad (e_{j-1},e_j)\in R,
    \label{eq:evidence-graph-path}
\end{equation}
where $V$ is the set of explored entities, $R$ is the set of relations supported by retrieved web evidence, and $h$ is the number of transitions in the exploration path. The resulting graph records the properties and relations collected for the entities along this process. It is then provided to a question-generation model, which constructs a multi-hop question--answer pair $(q,y^\ast)$ from the collected evidence. A subsequent difficulty-enhancement step removes overly explicit clues while preserving the reference answer. The resulting question is then solved by the teacher with search tools, producing a search trajectory whose final answer is judged before conversion to the training format.

Hereafter, $e_0$ is the \emph{seed entity}. A related entity considered during expansion is a \emph{candidate next-hop entity}; once selected, it becomes a \emph{next-hop entity} on $P$. These are the two entity-admission points at which SEF can intervene. We refer to synthesis without familiarity screening as \emph{ungated synthesis}, and to the variant that admits only entities reported as unfamiliar as \emph{SEF-gated synthesis}. The two variants are otherwise identical.

\subsection{Study Design}
\label{sec:study-design}

Building on the two entity-selection points described above, we organize our empirical study around three research questions (RQs):
\begin{enumerate}[label=\textbf{RQ\arabic*:},leftmargin=3.3em,itemsep=2pt,topsep=3pt]

\item \textbf{Signal.} Is familiarity with an entity $e$ predictive of teacher-relative difficulty, as reflected in question-level closed-book answerability?

\item \textbf{Intervention.} Does excluding entities that the teacher reports as familiar increase observed search demand and improve downstream student performance?

\item \textbf{Reuse and scaling.} Can previously collected evidence graphs be reused with a new teacher to expand the deep-search training data generated by that teacher?

\end{enumerate}

\section{Entity Familiarity as a Signal of Teacher-Relative Difficulty}
\label{sec:familiarity-signal}

This section investigates whether teacher-specific entity familiarity can provide an early signal of teacher-relative difficulty, addressing RQ1. We first introduce Evidence-Verified Entity Familiarity (EVF), an evidence-backed continuous estimator, and test whether questions with more familiar seed entities are easier for the teacher to answer without search. Based on the EVF analysis, we further introduce Self-Reported Entity Familiarity (SEF), a simpler binary self-report designed to serve a similar entity-screening role at substantially lower cost. Finally, we compare the operational costs of EVF and SEF to assess their suitability for repeated use during data synthesis.

\subsection{EVF Tracks Closed-Book Answerability}
\label{sec:familiarity-assessment}

We test whether a model's familiarity with a seed entity is associated with its teacher-relative difficulty on the corresponding question. We operationalize entity familiarity using Evidence-Verified Entity Familiarity (EVF) and teacher-relative difficulty using closed-book correctness. Closed-book accuracy is the mean correctness over a set of questions. Our analysis covers GLM-5.1, DeepSeek-V4-Flash, and Kimi-K3, with 300 questions evaluated for each model.

For each question, let $e_0$ denote its \emph{seed entity}, the first entity on the exploration path. Model $M_T$ first produces a factual brief about $e_0$ without tools. The same model, now operating with search access, retrieves external evidence and verifies the claims in the brief. Let $\mathcal{C}_{M_T}(e_0)$ denote the set of generated factual claims, and let $z(c)\in\{0,1\}$ indicate whether claim $c$ is verified by the retrieved evidence. We define EVF as
\begin{equation}
    \operatorname{EVF}_{M_T}(e_0)
    =
    \frac{1}{|\mathcal{C}_{M_T}(e_0)|}
    \sum_{c\in\mathcal{C}_{M_T}(e_0)} z(c).
    \label{eq:evf}
\end{equation}
EVF measures the fraction of the model-generated claims that can be verified against external evidence, with higher values indicating greater estimated familiarity. To measure teacher-relative difficulty, we ask the model to answer the corresponding question without tools and record whether its final answer is correct. Lower closed-book correctness indicates greater teacher-relative difficulty. Kimi-K3 judges final-answer correctness in all three model settings.

\begingroup
\setlength{\intextsep}{5pt plus 3pt minus 3pt}
\begin{table}[H]
    \centering
    \caption{Association between EVF and closed-book (CB) correctness. Failures are counted as incorrect. Low and high EVF denote $\operatorname{EVF}<0.25$ and $\operatorname{EVF}\in[0.75,1]$, respectively.}
    \label{tab:familiarity-signal}
    \small
    \setlength{\tabcolsep}{3.2pt}
    \begin{adjustbox}{max width=\textwidth}
    \begin{tabular}{lcccc}
        \toprule
        Teacher
        & EVF AUC
        & $\rho$ ($p$)
        & Low-EVF CB acc.
        & High-EVF CB acc. \\
        \midrule
        GLM-5.1
        & .600
        & .174 (.003)
        & 28.41\%
        & 45.63\% \\
        DeepSeek-V4-Flash
        & .620
        & .204 ($<$.001)
        & 26.67\%
        & 46.40\% \\
        Kimi-K3
        & .706
        & .351 ($<$.001)
        & 46.05\%
        & 80.98\% \\
        \bottomrule
    \end{tabular}
    \end{adjustbox}
\end{table}
\endgroup

Table~\ref{tab:familiarity-signal} shows that EVF is positively associated with closed-book correctness across all three teachers, with AUCs of .600--.706 and significant positive correlations. Questions with low-EVF seed entities are also consistently less answerable without tools than those with high-EVF seed entities.  We use $\operatorname{EVF}<0.25$ as a fixed operational definition of low familiarity; because EVF depends on both generated claims and retrieved evidence, we treat it as a diagnostic proxy rather than ground-truth model knowledge.

\subsection{SEF Provides a High-Recall Low-Familiarity Gate}
\label{sec:sef-validity}

EVF requires factual generation, web search, and evidence verification, making it too expensive for repeated use during graph construction. We therefore ask whether a simpler and less costly signal can retain most low-EVF entities while reducing the candidate pool before synthesis.

Our experiments reveal a useful pattern: within the low-EVF region, teachers often report being unfamiliar with the corresponding seed entity. This observation motivates a lightweight binary gate for graph construction, which we call Self-Reported Entity Familiarity (SEF):
\begin{equation}
    \operatorname{SEF}_{M_T}(e)
    \in
    \{\textsc{yes},\textsc{no}\},
    \label{eq:sef}
\end{equation}
where \textsc{yes} and \textsc{no} indicate that the teacher reports being familiar or unfamiliar with entity $e$, respectively. SEF requires one tool-free teacher call per entity. We compare closed-book accuracy by the seed entity's SEF judgment, then evaluate its screening behavior through low-EVF retention, high-EVF exclusion, and changes in pool composition.

\begingroup
\setlength{\intextsep}{5pt plus 3pt minus 3pt}
\begin{table}[H]
    \centering
    \caption{Closed-book (CB) accuracy by seed-entity SEF judgment and SEF screening of seed entities. Low and high EVF denote $\operatorname{EVF}<0.25$ and $\operatorname{EVF}\in[0.75,1]$, respectively. Retention is the fraction of low-EVF entities judged \textsc{no}; exclusion is the fraction of high-EVF entities judged \textsc{yes}. ``Before'' refers to all evaluated seed entities and ``after'' to the subset judged \textsc{no}.}
    \label{tab:sef-lowevf-gate}
    \small
    \setlength{\tabcolsep}{3.2pt}
    \resizebox{\textwidth}{!}{%
    \begin{tabular}{lrrrrrr}
        \toprule
        \multirow{2}{*}{Teacher}
        & \multicolumn{2}{c}{CB accuracy by seed SEF}
        & \multicolumn{4}{c}{SEF screening relative to EVF} \\
        \cmidrule(lr){2-3}\cmidrule(lr){4-7}
        & \textsc{yes}
        & \textsc{no}
        & \shortstack{Low-EVF\\retention (\%)}
        & \shortstack{High-EVF\\exclusion (\%)}
        & \shortstack{Low-EVF share (\%)\\Before $\rightarrow$ after}
        & \shortstack{High-EVF share (\%)\\Before $\rightarrow$ after} \\
        \midrule
        GLM-5.1
        & 44.06\%
        & 35.81\%
        & 96.59\%
        & 71.70\%
        & 30.14\% $\rightarrow$ 55.19\%
        & 54.45\% $\rightarrow$ 29.22\% \\
        DeepSeek-V4-Flash
        & 44.29\%
        & 27.04\%
        & 97.14\%
        & 81.60\%
        & 35.84\% $\rightarrow$ 65.38\%
        & 42.66\% $\rightarrow$ 14.74\% \\
        Kimi-K3
        & 84.24\%
        & 57.04\%
        & 94.74\%
        & 71.71\%
        & 26.21\% $\rightarrow$ 53.73\%
        & 70.69\% $\rightarrow$ 43.28\% \\
        \bottomrule
    \end{tabular}}
\end{table}
\endgroup

Table~\ref{tab:sef-lowevf-gate} shows that questions with \textsc{no} seed entities have lower closed-book accuracy across all three teachers. SEF also retains nearly all low-EVF seed entities while excluding most high-EVF entities. The retained pool has a higher low-EVF share and a lower high-EVF share, although a substantial high-EVF fraction remains for Kimi-K3. Although SEF is not a precise measure of teacher knowledge, it provides an informative signal of teacher-specific entity familiarity.

\subsection{SEF Enables Low-Cost Early Screening}
\label{sec:familiarity-cost}

Table~\ref{tab:familiarity-cost} compares the measured decision costs of SEF and EVF. SEF requires one tool-free LLM call per entity, whereas EVF must generate a factual brief, search, and verify evidence.

\begingroup
\setlength{\intextsep}{5pt plus 3pt minus 3pt}
\begin{table}[H]
    \centering
    \caption{Decision cost of familiarity assessment. Calls are mean LLM/tool calls per assessed entity, and wall-clock times are measured within each deployment setting.}
    \label{tab:familiarity-cost}
    \small
    \setlength{\tabcolsep}{5pt}
    \begin{adjustbox}{max width=\textwidth}
    \begin{tabular}{llccc}
        \toprule
        Teacher
        & Method
        & LLM calls
        & Tool calls
        & Mean wall time \\
        \midrule
        GLM-5.1
        & SEF
        & 1.00
        & 0.00
        & 5.28 s \\
        GLM-5.1
        & EVF
        & 10.86
        & 14.27
        & 107.83 s \\
        \midrule
        Kimi-K3
        & SEF
        & 1.00
        & 0.00
        & 24.32 s \\
        Kimi-K3
        & EVF
        & 8.35
        & 10.45
        & 224.07 s \\
        \bottomrule
    \end{tabular}
    \end{adjustbox}
\end{table}
\endgroup

Table~\ref{tab:familiarity-cost} shows that SEF requires far fewer LLM and tool calls than EVF, with mean wall-clock time lower by roughly $9$--$20\times$ in the measured deployment settings. This low per-decision cost makes repeated screening at entity-admission points practical during graph construction.

\begin{rqanswer}{RQ1 (Signal)}
Entity familiarity is an informative early signal of teacher-relative difficulty: questions built from less familiar entities are less likely to be answered without search. A single direct familiarity judgment provides a practical gate for selecting entities during entity-graph construction.
\end{rqanswer}

\section{SEF-Gated Synthesis and Distillation}
\label{sec:distillation}

\begin{figure}[t]
    \centering
    \includegraphics[width=\linewidth,pagebox=cropbox]{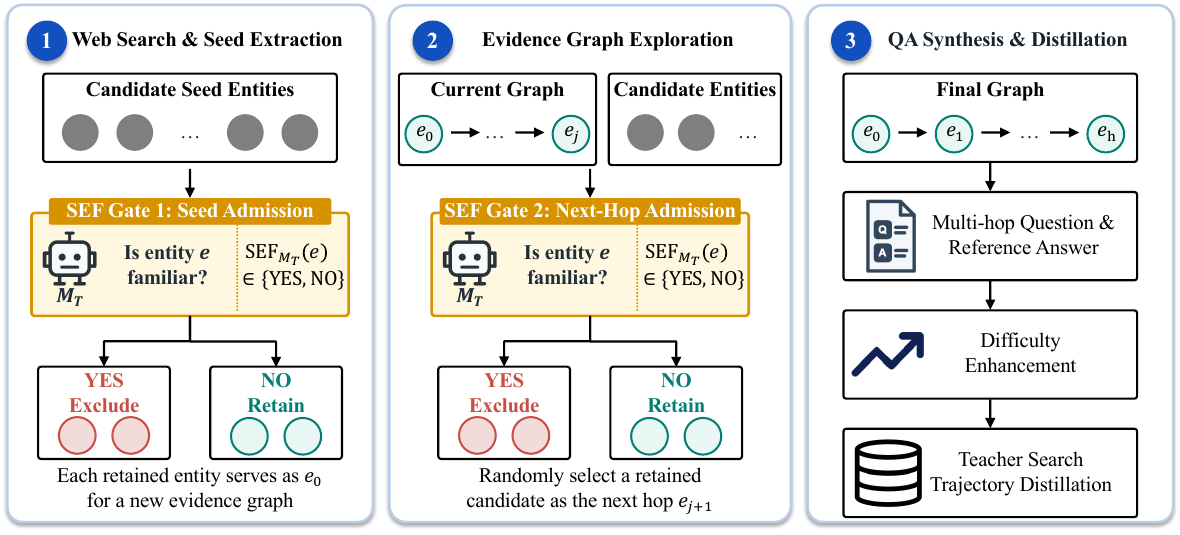}
    \vspace{-6pt}
    \caption{SEF-gated graph-based synthesis and trajectory collection. The target teacher $M_T$ screens candidate seed and next-hop entities, retaining only those it reports as unfamiliar ($\operatorname{SEF}_{M_T}(e)=\text{\textsc{no}}$). Retained seeds initialize evidence graphs, and next hops are sampled from retained candidates. The completed graph supports multi-hop question--answer generation and difficulty enhancement, followed by tool-assisted teacher search to collect trajectories for distillation.}
    \label{fig:sef-gated-synthesis}
    \vspace{-6pt}
\end{figure}

We evaluate whether using SEF to admit entities changes the teacher's search behavior and produces more useful training data. Figure~\ref{fig:sef-gated-synthesis} summarizes the pipeline: the target teacher screens candidate seeds before graph construction and candidate next-hop entities during graph expansion, while the downstream question-generation, difficulty-enhancement, and teacher-search stages remain unchanged from ungated synthesis. We then measure search demand in teacher trajectories and compare student performance under equal-sized training sets.

\subsection{SEF-Gated Entity Admission}
\label{sec:sef-synthesis}

Section~\ref{sec:familiarity-signal} evaluates EVF and SEF only on seed entities $e_0$. Nevertheless, a multi-hop question also draws on later entities along its exploration path. An unfamiliar seed entity alone therefore does not ensure that all entities supporting the question are unfamiliar to the teacher. Accordingly, SEF-gated synthesis applies SEF at both entity-admission points: before graph construction for seed entities and during expansion for candidate next-hop entities.

At expansion step $k$, let $\mathcal{C}_k$ be the candidate next-hop entities. Before sampling the next-hop entity, we form an admissible set containing only those that the target teacher $M_T$ reports as unfamiliar:
\begin{equation}
  \mathcal{C}^{-}_k =
  \{e\in\mathcal{C}_k : \operatorname{SEF}_{M_T}(e)=\text{\textsc{no}}\},
  \qquad
  e_{k+1}\sim\operatorname{Uniform}(\mathcal{C}^{-}_k).
  \label{eq:sef-filter}
\end{equation}
If $\mathcal{C}^{-}_k$ is empty, expansion of that branch stops and the partial graph is preserved. A teacher-specific cache reuses earlier SEF judgments whenever an entity is encountered again.

Ungated and SEF-gated synthesis follow the same graph-based pipeline described in Section~\ref{sec:synthesis-pipeline}; they differ only in entity admission.

\subsection{SEF Gating Increases Observed Search Demand}
\label{sec:teacher-effects}

We compare ungated and SEF-gated synthesis for GLM-5.1, DeepSeek-V4-Flash, and Kimi-K3, yielding three teacher-specific comparisons and six trajectory pools. Each pool contains more than 2,000 trajectories. For DeepSeek-V4-Flash and Kimi-K3, whose reasoning effort can be configured, we set \texttt{reasoning\_effort} to \texttt{high}. Kimi-K3 judges whether each trajectory's final answer is correct. For each full trajectory pool, we report mean LLM turns, reasoning tokens, and structured tool calls per trajectory, along with tool-assisted solve accuracy.

\begingroup
\setlength{\intextsep}{6pt plus 2pt minus 2pt}
\begin{table}[H]
    \centering
    \caption{Teacher trajectory statistics for ungated $\rightarrow$ SEF-gated synthesis. All per-trajectory means and solve accuracies use the complete raw trajectory pools.}
    \label{tab:teacher-trajectories}
    \small
    \setlength{\tabcolsep}{4.5pt}
    \begin{adjustbox}{max width=\textwidth}
    \begin{tabular}{lrrrr}
        \toprule
        Teacher & LLM turns & Reasoning tokens & Tool calls & Solve acc. (\%) \\
        \midrule
        GLM-5.1 & $24.01\!\rightarrow\!32.32$ & $19{,}587\!\rightarrow\!23{,}169$ & $35.05\!\rightarrow\!48.72$ & $76.89\!\rightarrow\!64.87$ \\
        DeepSeek-V4-Flash & $39.20\!\rightarrow\!40.12$ & $20{,}846\!\rightarrow\!21{,}857$ & $43.78\!\rightarrow\!45.05$ & $76.96\!\rightarrow\!71.46$ \\
        Kimi-K3 & $25.50\!\rightarrow\!31.00$ & $4{,}700\!\rightarrow\!5{,}211$ & $65.04\!\rightarrow\!77.66$ & $96.35\!\rightarrow\!95.43$ \\
        \bottomrule
    \end{tabular}
    \end{adjustbox}
\end{table}
\endgroup

Table~\ref{tab:teacher-trajectories} reports ungated $\rightarrow$ SEF-gated results on the full teacher pools. All nine search-demand measures increase: LLM turns by 2.3--34.6\%, reasoning tokens by 4.8--18.3\%, and tool calls by 2.9--39.0\%. Tool-assisted solve accuracy falls by 12.02 and 5.50 percentage points for GLM-5.1 and DeepSeek-V4-Flash, respectively. Kimi-K3 makes 21.6\% more turns and 19.4\% more tool calls, while its accuracy changes by only $-0.92$ points, showing that greater search demand can emerge even when a strong teacher largely maintains its success rate. Together, these differences are consistent with greater teacher-relative difficulty for questions from SEF-gated synthesis.

\subsection{SEF-Gated Trajectories Yield Higher Student Accuracy}
\label{sec:student-effects}

The teacher-side statistics characterize search demand but do not determine the training value of the resulting data. We therefore evaluate downstream utility separately under equal-sized training sets, holding the number of trajectories fixed within each teacher setting. For GLM-5.1 and DeepSeek-V4-Flash, we train Qwen3-30B-A3B-Thinking on 1,490 trajectories from each synthesis variant per teacher. For Kimi-K3, we train Qwen3.8-27B on 2,200 trajectories from each synthesis variant. We evaluate all students on the 130-question BrowseComp subset selected by QUEST~\citep{quest2026}. Each condition is run independently three times with a maximum context length of 256k, and we report the mean accuracy and sample standard deviation across runs. We use the same tool configuration and tool-call definitions as OpenResearcher and use Kimi-K3 to judge answer correctness. \textit{To reduce benchmark-answer leakage, the search tools filter Hugging Face URLs from search results and block direct access to those pages.}

\begingroup
\setlength{\intextsep}{6pt plus 2pt minus 2pt}
\begin{table}[H]
    \centering
    \caption{Student accuracy on the 130-question BrowseComp subset selected by QUEST. Except for the approximate Qwen3-30B-A3B-Thinking baseline, values are mean $\pm$ sample standard deviation over three independent runs.}
    \label{tab:student-accuracy}
    \small
    \setlength{\tabcolsep}{4.5pt}
    \begin{adjustbox}{max width=\textwidth}
    \begin{tabular}{llrrrr}
        \toprule
        Trajectory teacher & Student & Before SFT & Ungated & SEF-gated & Gain (pp) \\
        \midrule
        GLM-5.1 & Qwen3-30B-A3B-Thinking & $\sim$2.5\% & $33.08\pm0.77$\% & $41.28\pm1.94$\% & $+8.20$ \\
        DeepSeek-V4-Flash & Qwen3-30B-A3B-Thinking & $\sim$2.5\% & $25.90\pm2.91$\% & $32.05\pm1.18$\% & $+6.15$ \\
        Kimi-K3 & Qwen3.8-27B & $44.10\pm2.47$\% & $55.13\pm4.37$\% & $63.59\pm1.18$\% & $+8.46$ \\
        \bottomrule
    \end{tabular}
    \end{adjustbox}
\end{table}
\endgroup

Table~\ref{tab:student-accuracy} shows that both synthesis variants improve mean accuracy over the corresponding before-SFT student. With the number of training trajectories held fixed, SEF-gated trajectories yield mean accuracy gains over ungated trajectories of 8.20, 6.15, and 8.46 percentage points for GLM-5.1, DeepSeek-V4-Flash, and Kimi-K3, respectively. These results favor SEF gating across three trajectory teachers and two student backbones. Together with the teacher-side results, they indicate greater search demand during synthesis and stronger student performance when trained on the same number of trajectories.

\begin{rqanswer}{RQ2 (Intervention)}
Across the evaluated settings, excluding entities already familiar to the teacher yields higher observed search demand, while equal-sized SEF-gated trajectory sets train students with higher accuracy than the corresponding ungated sets. These parallel outcomes support entity familiarity as a practical control for data synthesis.
\end{rqanswer}

\section{Cross-Teacher Adaptation through Graph Reuse}
\label{sec:graph-reuse}

SEF judgments are teacher-specific, raising a practical concern: adapting to each new teacher by exploring evidence graphs from scratch with SEF gating would repeat the costly work of entity discovery and evidence collection. In this section, we therefore investigate whether existing evidence graphs can be reused. 

\subsection{SEF Judgments Overlap but Require Target-Specific Reassessment}
\label{sec:cross-teacher-agreement}

\begingroup
\setlength{\intextsep}{5pt plus 3pt minus 3pt}
\begin{table}[H]
\centering
\caption{Pairwise SEF judgments on the shared 5,995-entity pool.}
\label{tab:sef-agreement}
\small
\setlength{\tabcolsep}{4pt}
\begin{adjustbox}{max width=\linewidth}
\begin{tabular}{lrrrr}
\toprule
Teacher pair ($A$ / $B$) & \shortstack{Both\\unfamiliar} & \shortstack{$A$ familiar\\$B$ unfamiliar} & \shortstack{$A$ unfamiliar\\$B$ familiar} & Agreement \\
\midrule
GLM-5.1 / DeepSeek-V4-Flash & 29.6\% & 9.6\% & 6.0\% & 84.4\% \\
GLM-5.1 / Kimi-K3 & 32.8\% & 7.3\% & 2.8\% & 89.9\% \\
DeepSeek-V4-Flash / Kimi-K3 & 32.2\% & 8.0\% & 7.2\% & 84.8\% \\
\bottomrule
\end{tabular}
\end{adjustbox}
\end{table}
\endgroup

Table~\ref{tab:sef-agreement} compares SEF judgments by GLM-5.1, DeepSeek-V4-Flash, and Kimi-K3 on entities sampled from three teacher-specific pools (2,000 per pool). Pairwise agreement is high (84.4--89.9\%), but directional disagreements still account for 10.1--15.6\% across teacher pairs. These differences mean that a source teacher's SEF decision cannot be transferred directly: the target teacher must reassess candidate paths before reuse.

\subsection{Reuse Evidence Graphs, Regenerate Target-Specific Supervision}
\label{sec:reuse-method}

We reuse only the evidence graph---including its entities, evidence-backed relations, retrieved evidence, and candidate paths---rather than the source teacher's trajectories. For a candidate path $P=(e_0,\ldots,e_h)$ and target teacher $M_T$, we accept the path only if all of its entities are judged unfamiliar:
\begin{equation}
\operatorname{Accept}_{M_T}(P)
=
\mathbf{1}\!\left[
\forall e\in P,\;
\operatorname{SEF}_{M_T}(e)=\text{\textsc{no}}
\right].
\label{eq:reuse-path}
\end{equation}
Accepted paths are then used to resynthesize question--answer pairs and collect new search trajectories from the target teacher. Graph reuse therefore preserves prior entity-discovery and evidence-collection work, while familiarity screening, question synthesis, and trajectory generation remain target-specific.

\subsection{Scaling Target-Specific Supervision with Reused Graphs}
\label{sec:reuse-scaling}

We use Kimi-K3 as the target teacher. Starting from the 2,200 SEF-gated trajectories of Kimi-K3 in Section~\ref{sec:distillation}, we load 7,143 evidence graphs collected for the other teachers and reapply SEF with Kimi-K3 to their candidate paths. Only paths whose entities pass the Kimi-K3-specific SEF criterion in Eq.~\ref{eq:reuse-path} are retained for question resynthesis and trajectory generation. This target-specific screening yields 5,411 additional Kimi-K3 trajectories, expanding the training set to 7,611 trajectories.

We evaluate Qwen3.8-27B on the 130-question BrowseComp subset selected by QUEST~\citep{quest2026} with a 256k context, reporting mean $\pm$ sample standard deviation over three independent runs. With 2,200 native SEF-gated trajectories, the student reaches $63.59\pm1.18$\% accuracy; adding 5,411 trajectories generated from reused graphs raises accuracy to $66.41\pm2.35$\%, a 2.82-percentage-point gain. This higher mean suggests downstream value from expanding training data through graph reuse, without isolating reuse from the effect of additional training examples.

\begin{rqanswer}{RQ3 (Reuse and Scaling)}
Evidence graphs can transfer across teachers after target-specific SEF reassessment, expanding training data. In this data-expansion setting, the additional trajectories are associated with higher mean student accuracy.
\end{rqanswer}

Table~\ref{tab:scaling-comparison} summarizes results on BrowseComp (BC), BrowseComp-ZH (BC-ZH)~\citep{browsecompzh}, SEAL-0~\citep{sealqa}, the text-only subset of Humanity's Last Exam with tools~\citep{hle}, and xbench-DeepSearch-2510~\citep{xbench2510}. We evaluate on the full evaluation set of each benchmark, using the text-only subset for HLE. For BC and BC-ZH, we report results both with a ReAct setup and with context management.

\begingroup
\setlength{\intextsep}{5pt plus 3pt minus 3pt}
\begin{table}[H]
\centering
\caption{Comparison of deep research agents across five benchmarks.}
\label{tab:scaling-comparison}
\small
\setlength{\tabcolsep}{3pt}
\begin{tabularx}{\linewidth}{@{}>{\raggedright\arraybackslash}Xccccc@{}}
\toprule
Method & BC & BC-ZH & SEAL-0 & \shortstack{HLE text\\+ tools} & \shortstack{xbench\\2510} \\
\midrule
OpenAI Deep Research~\citep{browsecomp,browsecompzh} & 51.5 & 42.9 & -- & -- & -- \\
Tongyi DeepResearch 30B-A3B~\citep{tongyi_dr} & $43.4/58.3^{*}$ & $46.7/58.1^{*}$ & -- & $32.9^{\dagger}/38.3^{*\dagger}$ & 55.0 \\
OffSeeker-8B-DPO~\citep{offseeker} & 12.8 & 26.6 & -- & 13.8 & -- \\
OpenResearcher-30B-A3B~\citep{openresearcher2026} & 26.3 & -- & -- & $19.6^{*}$ & -- \\
QUEST-35B~\citep{quest2026} & $45.5^{*\dagger}/64.6^{*}$ & -- & -- & $37.2^{*}$ & -- \\
Apodex-1.0-mini~\citep{apodex2026} & $71.5^{*}$ & $\textbf{80.6}^{*}$ & -- & 46.8 & -- \\
FORT-Searcher~\citep{fort2026} & $55.9/72.2^{*}$ & $62.1/75.0^{*}$ & $43.7/46.0^{*}$ & -- & $54.1/57.2^{*}$ \\
DeepSearch-World-9B~\citep{deepsearchworld2026} & $31.2^{*}$ & $36.4^{*}$ & -- & $25.7^{*}$ & -- \\
AREX-Turbo~\citep{arex2026} & $70.7^{*}$ & -- & -- & $40.6^{*}$ & $57.0^{*}$ \\
\midrule
Ours (Qwen3.8-27B) & $66.2/\textbf{74.3}^{*}$ & $64.7/69.6^{*}$ & \textbf{49.5} & \textbf{48.6} & \textbf{65.0} \\
\bottomrule
\end{tabularx}
\end{table}
\endgroup
\vspace{-6pt}

$^{*}$ marks a result with explicit context management, $^{\dagger}$ marks an evaluation on fewer examples than the target split, and a dash denotes an unavailable or incompatible result. Our method performs competitively across multiple benchmarks. However, student models and training procedures differ across methods; some use mid-training and reinforcement learning rather than supervised fine-tuning alone. The comparisons should therefore be interpreted as indicative rather than strictly controlled.

\section{Related Work}
\label{sec:related-work}

\paragraph{Synthetic Deep-Search Data and Difficulty Control.}
Two mechanisms are often used to control the difficulty of synthetic deep-search questions, sometimes in combination. One shapes entity paths, reasoning constraints, evidence distribution, or clues \citep{websailor,opendatasynthesis,deepresearch9k,webshaper,tan2026searchmaster}; TaskCraft expands task depth and breadth in a broader agent setting \citep{taskcraft}, while FORT-Searcher selects long-tail graph roots to reduce prior-knowledge shortcuts \citep{fort2026}. The other uses solver feedback to revise or filter candidates \citep{sage,pandit2025agentic,drzero,openseeker}; OpenSeeker discards questions answered correctly without tools. WebLeaper combines structural task design with trajectory filtering \citep{webleaper}. We study a distinct but compatible, teacher-relative dimension: how the intended teacher's familiarity with selected entities relates to the difficulty of the resulting questions. Building on an exsiting graph-based pipeline \citep{offseeker}, SEF screens seeds and candidate next-hop entities using the target teacher's self-report before question generation.

\paragraph{Are Search Agents Actually Searching?}
Closed-book, evidence-blocking, and counterfactual interventions test whether answers persist without usable evidence, exposing memory-backed verification as well as over- and under-searching \citep{livebrowsecomp,zhang2026decision,kim2026whensearch}.
Separate audits show that web agents can retrieve benchmark copies or answer keys at inference time and thereby inflate reported scores, while dynamic, recent, and counterfactual benchmarks seek to reduce contamination and parametric-memory shortcuts \citep{han2025searchtime,wang2026searchtime,livebrowsecomp,wang2026evobrowsecomp,chen2026paraworld}.
Distillation evidence also warns that fluent imitation need not transfer factual capability \citep{gudibande2023false}; related agent-distillation pipelines filter correct traces, align supervision with student capacity, or preserve only searches that are counterfactually necessary \citep{kang2025agentdistill,tang2026smartad,cai2026thinkbig}. 
Prior work thus treats search versus recall chiefly as an evaluation behavior to diagnose or repair on already-formed questions and trajectories \citep{liang2026searchauditor}.
We instead make it an entity-level data-construction rule, and evaluate consequences for search demand in teacher trajectories and downstream student gains.

\section{Conclusion}
\label{sec:conclusion}

In this paper, we investigated how teacher-specific entity familiarity can guide the synthesis of deep-search distillation data. We first introduced EVF to assess familiarity against web evidence and found that it is associated with teachers' closed-book answerability, although its verification cost limits its use during graph construction. We therefore used the lighter-weight SEF judgment to screen seed and next-hop entities. Across three teachers, SEF-gated synthesis produced trajectories with greater observed search demand and yielded higher student accuracy than ungated synthesis in equal-sized training comparisons. We further found that target-specific SEF reassessment enables cross-teacher graph reuse: expanding Kimi-K3's training set with reused graphs achieved a further gain in student accuracy. Together, these results support using teacher familiarity as an early data-construction signal rather than relying on graph structure alone to anticipate useful difficulty.

\paragraph{Limitations.}
EVF and SEF are operational signals of teacher behavior rather than ground-truth or calibrated knowledge boundaries, and validity tests mainly probe root entities. Our empirical scope is limited to three teachers, two student backbones, and one fact-seeking evaluation set. The current method applies only to graph-based QA synthesis in which nodes represent entities; adapting entity-familiarity gating to other QA synthesis paradigms remains future work.

\bibliographystyle{assets/plainnat}
\bibliography{Bibs/iclr2027_conference}

\clearpage
\clearpage
\appendix
\section*{Appendix}

\section{Question Synthesis and Graph-Depth Diagnostics}
\label{app:synthesis-depth}

\subsection{DeepForge Question Synthesis}
\label{app:synthesis-flow}

We adapt the graph-based question-synthesis workflow of OffSeeker~\citep{offseeker}. Topic-diverse web pages are collected and used to extract seed entities. Starting with a seed $e_0$, the explorer searches and reads web evidence, records factual properties and relations, and proposes related entities. It samples a candidate next-hop entity from the available relations and repeats exploration up to a sampled depth, retaining a partial graph if the process stops early. Thus, the recorded transition path $P=(e_0,\ldots,e_h)$ describes exploration, not necessarily the reasoning steps required to answer a generated question.

The resulting evidence graph is passed to a question generator to produce a question and reference answer. A subsequent enhancement step removes overly explicit clues while keeping the answer fixed. The search-enabled teacher then attempts the enhanced question, and its final answer is judged before the trajectory is converted into student training data. In ungated synthesis, seed and next-hop entities are admitted without a familiarity test. SEF-gated synthesis instead screens both admission points with the target teacher's tool-free self-report and samples the next hop only from entities judged unfamiliar. If none passes, exploration stops without discarding the partial graph. These are entity-admission decisions made \emph{before} the downstream question and trajectory exist; no source-teacher question or trajectory is transferred when an evidence graph is later reused by a different teacher.

\subsection{Depth versus Closed-Book Answerability}
\label{app:depth-diagnostic}

We use the length of the recorded entity transition path as graph depth. Table~\ref{tab:app-depth-closedbook} groups each teacher's closed-book correctness by this depth. Missing final answers, length cutoffs, and timeouts count as incorrect.

\begin{table}[H]
\centering
\caption{Closed-book correctness by recorded graph depth. Cells give correct answers / questions and the corresponding accuracy; $\rho$ is the question-level Spearman correlation with its unadjusted $p$ value.}
\label{tab:app-depth-closedbook}
\small
\setlength{\tabcolsep}{3.5pt}
\begin{adjustbox}{max width=\linewidth}
\begin{tabular}{lrrrrr}
\toprule
Teacher & Depth 1 & Depth 2 & Depth 3 & Depth 4 & $\rho$ ($p$) \\
\midrule
GLM-5.1 & 9/28 (32.1\%) & 46/123 (37.4\%) & 49/120 (40.8\%) & 15/29 (51.7\%) & $+.086$ (.138) \\
DeepSeek-V4-Flash & 9/38 (23.7\%) & 43/124 (34.7\%) & 45/110 (40.9\%) & 8/28 (28.6\%) & $+.067$ (.250) \\
Kimi-K3 & 28/37 (75.7\%) & 82/112 (73.2\%) & 85/122 (69.7\%) & 21/29 (72.4\%) & $-.038$ (.517) \\
\bottomrule
\end{tabular}
\end{adjustbox}
\end{table}

GLM-5.1's binned accuracy rises with depth; DeepSeek-V4-Flash is nonmonotonic, and Kimi-K3 does not show an increasing trend. Hence these data do \emph{not} show that simply increasing exploration depth reliably decreases a teacher's closed-book answerability. They illustrate why graph-structural complexity need not translate into teacher-relative difficulty.

One possible explanation for GLM-5.1's increasing accuracy with depth is that longer paths provide more opportunities to encounter well-known entities. If prominent entities are proposed more often as next-hop candidates, an explorer selecting among the proposed relations may reach familiar entities even when the seed entity is obscure. The present analysis does not measure entity connectivity or familiarity at each hop, so this mechanism remains a hypothesis.

\section{Exact EVF and SEF Prompt Templates}
\label{app:familiarity-prompts}

We provide the exact prompts used for EVF and SEF. Entity names and factual briefs are inserted into the templates at inference time. SEF and EVF brief generation use \emph{user} messages without an explicit system message; EVF verification uses both a system message and an entity-specific user message. All calls use temperature zero. The EVF verifier is the probed teacher with browser access, whereas Kimi-K3 judges the final answers to synthesized questions.

\subsection{SEF Self-Report}
\label{app:sef-prompt}

The one-call, tool-free self-report uses the following user-message template:
\begingroup\scriptsize
\begin{verbatim}
Do you have enough specific factual knowledge about the entity named "{name}"
to answer detailed questions about its history, properties, or attributes?

Reply with only one word: YES or NO.
- YES: you can recall multiple specific verifiable facts about this exact entity
- NO: you cannot, or you're unsure which entity this refers to, or your knowledge
  is vague/category-level only

If in doubt, answer NO.
\end{verbatim}
\endgroup

During synthesis, judgments are cached by entity and teacher-model identifier; the signal and cost evaluations make uncached calls.

\subsection{EVF Factual Brief}
\label{app:evf-brief-prompt}

The teacher produces its factual brief without browser tools, using this user-message template:
\begingroup\scriptsize
\begin{verbatim}
Tell me everything you confidently know about the entity named: {name}

Write a single paragraph of factual statements. Each statement should be a
verifiable claim such as a date, place, role, relationship, organization,
work, achievement, or attribute.

Rules:
- Write only facts you are confident about. Do not speculate or fill space.
- Do NOT use hedging language like "I think", "possibly", "around",
  "approximately", "may have", "is believed to be". State each fact directly
  or omit it.
- Do NOT add disclaimers about your knowledge cutoff or your uncertainty.
- Plain prose only. No markdown, no bullet lists, no headings.
- Keep it under 150 words. Quality over quantity.
- If you have no reliable information about this entity, output exactly:
  I have no information about this entity.
\end{verbatim}
\endgroup

When the teacher provides no factual brief, EVF is set to zero without evidence verification.

\subsection{EVF Evidence Verification}
\label{app:evf-verification-prompt}

The same teacher then verifies the brief using browser search, open, and find tools. The verification system message is:
\begingroup\fontsize{6.3}{7.5}\selectfont
\begin{verbatim}
You are a fact-checking assistant with access to browser tools
(browser.search, browser.open, browser.find). You will be given an entity
name and a candidate brief about that entity, written by another LLM with
no tool access. Your job is to extract atomic factual claims from the brief
and verify each via the browser tools.

Procedure:
1. Read the brief and extract up to 8 atomic factual claims. Each claim must:
   - Be a single self-contained proposition (no compound statements joined
     by "and" / "but" / ";").
   - Use the entity's full name rather than pronouns (resolve "he/she/it"
     to the entity).
   - Be a concrete fact in principle verifiable through web search.
   - Not be a trivial restatement of the entity's name or category.
2. For each claim, decide its verdict by actually issuing browser searches:
   - "true"    : confirmed by at least one reliable source you retrieved
                 in this conversation.
   - "false"   : contradicted by a reliable source you retrieved.
   - "unknown" : you searched but could not find evidence either way, or
                 sources disagree, or you could not pin the claim down to
                 the precision stated in the brief.
3. Be strict. Hedging or partial matches are NOT "true". If the brief says
    "founded in 1850" and your evidence says "founded in the 1850s", that
\end{verbatim}\vspace{-0.8\baselineskip}
\noindent\hspace*{2.4em}{\ttfamily is "unknown" }{\rmfamily\textemdash}{\ttfamily\ the precise year is not confirmed.}\par
\begin{verbatim}

When verification is complete, output your final answer as JSON inside
<result>...</result> tags, with no text after the closing tag:

<result>
{
  "claims": [
    {"text": "<atomic claim>", "verdict": "true|false|unknown", "evidence": "<short URL or quote>"}
  ]
}
</result>

If the candidate brief is exactly the sentinel
"I have no information about this entity." (or trivially close to it),
skip all searches and output:

<result>{"claims": []}</result>
\end{verbatim}
\endgroup

The accompanying user message has the following form:
\begingroup\scriptsize
\begin{verbatim}
Entity: {entity_name}

Candidate brief:
{brief}
\end{verbatim}
\endgroup
The verifier prompt refers to the brief writer as \emph{another LLM}, although the same teacher model fills both roles in these experiments, first without and then with browser access. Each claim labeled \texttt{true} contributes one to the verified-claim count; \texttt{false} and \texttt{unknown} do not. EVF divides the verified-claim count by the number of extracted claims and assigns zero if the list is empty.

\subsection{Verifier Turn Limit}
\label{app:evf-forced-prompt}

If verification reaches its configured tool-loop limit, the teacher receives the following \emph{user} message and must return a final answer without further tool calls:
\begingroup
\begin{quote}\small\fontencoding{T1}\ttfamily\selectfont
STOP. Do not use any more tools. Based only on the searches you have already performed in this conversation, output the result JSON now. Mark any claim you could not confirm as "unknown". Output only the <result>...</result> block and nothing else.
\end{quote}
\endgroup

The turn limit is configurable. Results obtained after forced finalization are not stored in the EVF cache. Cost measurements were performed without cache reuse.

\section{Evaluation Details for the Reported Tables}

\label{app:evaluation-details}

The analyses of teacher answers count length cutoffs, timeouts, empty final answers, and missing judgments as incorrect. Answered-only accuracy is used solely to diagnose generation failures. Kimi-K3 judges the final answers from both the closed-book evaluations and the teachers' tool-assisted search trajectories. EVF claim verification instead uses the probed teacher with browser access (Appendix~\ref{app:evf-verification-prompt}). The three familiarity samples and the ungated/gated training-data pools are teacher-specific.

\subsection{Table~\ref{tab:familiarity-signal}: EVF and Closed-Book Solving}

\label{app:eval-evf}

For GLM-5.1, DeepSeek-V4-Flash, and Kimi-K3, the familiarity analysis samples enhanced questions from teacher-specific QA pools constructed using the same graph-based synthesis procedure. Each question is paired with its graph's seed entity. The teacher writes a tool-free factual brief and verifies its extracted claims with browser access using the prompts in Appendix~\ref{app:familiarity-prompts}. Unsuccessful EVF probes have no score, whereas a valid no-information brief receives zero.

Each teacher independently answers its own question without tools, and Kimi-K3 evaluates the final answer. EVF AUC binarizes each entity's mean closed-book correctness at $\geq .5$, while Spearman's $\rho$ uses the continuous entity-level outcome. The low and high EVF regions use the fixed cutoffs $[0,.25)$ and $[.75,1]$. Table~\ref{tab:app-depth-closedbook} stratifies the same closed-book outcomes by recorded graph depth.

\subsection{Table~\ref{tab:sef-lowevf-gate}: SEF Screening}

\label{app:eval-sef}

Each teacher gives a tool-free YES/NO self-report on the seed entity using Appendix~\ref{app:sef-prompt}. The analysis joins valid SEF and EVF outcomes by entity identity, with EVF$<.25$ defining the low-EVF region. Low-EVF retention and high-EVF exclusion measure screening within their respective EVF regions. Before/after shares measure the change in pool composition after retaining only \textsc{no} judgments; precision and enrichment characterize the low-EVF content of the retained pool. EVF is an operational proxy, not an externally verified knowledge label. The SEF labels themselves receive no external judgment.

\subsection{Table~\ref{tab:familiarity-cost}: Decision Costs}

\label{app:eval-cost}

EVF and SEF are evaluated without cache hits on teacher-specific entity samples. Table~\ref{tab:familiarity-cost} reports mean LLM calls, browser tool calls, and wall time per assessed entity. EVF includes the tool-free brief, every verifier turn, and any forced final call; SEF consists of one tool-free call. Wall times are comparable only within a deployment.

\subsection{Table~\ref{tab:teacher-trajectories}: Teacher Search Behavior}

\label{app:eval-teachers}

For each teacher, ungated and SEF-gated collections follow the same graph-based QA synthesis and tool-assisted solving stages. The gated condition adds SEF screening when admitting seed and next-hop entities. Each condition generates its own graph and question pool, so the two conditions do not contain paired questions or paths.

Turns, reasoning tokens, tool calls, and solve accuracy are measured on every raw teacher trajectory in each ungated or SEF-gated pool. Assistant messages count as turns, and structured browser actions count as tool calls. Nonempty assistant reasoning fields are tokenized with the Qwen3-30B-A3B-Thinking-2507 tokenizer; this proxy excludes tool observations and final answers and is not any teacher's native API token count. Kimi-K3 judges final-answer correctness, with empty predictions counted as incorrect. Because the ungated and gated graph and question pools are not paired, these comparisons do not isolate the effect of entity screening.

\subsection{Table~\ref{tab:student-accuracy}: Student Evaluation}
\label{app:eval-students}

Students are evaluated on the 130-question BrowseComp subset selected by QUEST~\citep{quest2026} with a 256k maximum context length. Each condition is run independently three times on all 130 questions. Kimi-K3 judges final-answer correctness, and generation failures, empty answers, and missing judgments count as incorrect. We report mean accuracy and the sample standard deviation across the three runs. The before-SFT Qwen3-30B-A3B-Thinking result is retained as an approximate $\sim$2.5\% reference; the other student results use the three-run aggregation.

To reduce benchmark-answer leakage during web evaluation, the browser removes search results whose URLs point to Hugging Face before their titles, snippets, or links are shown to the student, and it rejects direct attempts to open those pages. This filtering is applied in every reported student run.

\subsection{Table~\ref{tab:sef-agreement}: Cross-Teacher SEF Agreement}
\label{app:eval-agreement}

Each teacher contributes 2,000 sampled entities from its historical SEF cache. Deduplication produces a shared pool of 5,995 names, on which GLM-5.1, DeepSeek-V4-Flash, and Kimi-K3 each issue fresh tool-free SEF judgments. Table~\ref{tab:sef-agreement} reports the share judged unfamiliar by both teachers, the two directional disagreement shares, and overall pairwise agreement on valid judgments.

\subsection{Graph Reuse and Student Evaluation}
\label{app:eval-reuse}

Previously collected non-Kimi evidence graphs supply candidate paths for the Kimi-K3 target, but neither questions nor teacher trajectories. Kimi-K3 applies SEF to the candidate paths; accepted paths are used to resynthesize questions and collect new Kimi-K3 search trajectories. The 2,200 native trajectories and 5,411 trajectories from reused graphs form the 7,611-example training set.

Under the same 256k-context student-evaluation protocol, the model trained on the 2,200 native SEF-gated trajectories achieves 63.59$\pm$1.18\% accuracy, while the model trained on all 7,611 trajectories achieves 66.41$\pm$2.35\%. Each value is the mean and sample standard deviation over three independent runs on the QUEST-selected BrowseComp subset, with the failure and answer-judgment rules described in Appendix~\ref{app:eval-students}.

\section{Student Training Details}
\label{app:training-details}

The students are trained by supervised fine-tuning on serialized teacher search trajectories, not on the closed-book familiarity probes. Each six-epoch run uses full-parameter updates in bfloat16 with DeepSpeed ZeRO-3, a 262,144-token maximum sequence length, sequence packing, Adam betas $(.9,.95)$, weight decay $.1$, and gradient clipping at $1.0$. The student tokenizer and chat template serialize the complete sequence, including tool interactions. Both students use an effective global batch size of 32 packed sequences and a cosine learning-rate schedule. Qwen3-30B-A3B-Thinking uses a learning rate of $5\times10^{-5}$; Qwen3.8-27B uses $5\times10^{-6}$.

We convert judged teacher trajectories into SFT examples after removing records with no assistant messages, no tool use, malformed tool calls, or an unfinished reasoning or tool action at the end. Records that cannot be rendered in the student's conversation format are also removed. Incorrect answers are not excluded.

\end{document}